\documentclass[letterpaper]{article} 
\usepackage{aaai23}  
\usepackage{times}  
\usepackage{helvet}  
\usepackage{courier}  
\usepackage[hyphens]{url}  
\usepackage{graphicx} 
\usepackage{natbib}  
\usepackage{caption} 
\usepackage{xcolor}

\usepackage{algorithm}
\usepackage{algorithmic}
\usepackage{amsmath,amssymb,amsfonts}
\usepackage{newfloat}
\usepackage{listings}
\DeclareCaptionStyle{ruled}{labelfont=normalfont,labelsep=colon,strut=off} 
\floatstyle{ruled}
\newfloat{listing}{tb}{lst}{}
\floatname{listing}{Listing}
\title{YOLOV: Making Still Image Object Detectors Great at Video Object Detection}
\author {
    Yuheng Shi\textsuperscript{\rm 1},
    Naiyan Wang\textsuperscript{\rm 2},
    Xiaojie Guo\textsuperscript{\rm 1}\thanks{Corresponding author}
}
\affiliations {
    \textsuperscript{\rm 1} Tianjin University\\
    \textsuperscript{\rm 2} TuSimple\\
    yuheng@tju.edu.cn, winsty@gmail.com, xj.max.guo@gmail.com
}

\usepackage{bibentry}

\begin{document}

\maketitle

\begin{abstract}
Video object detection (VID) is challenging because of the high variation of object appearance as well as the diverse deterioration in some frames. On the positive side, the detection in a certain frame of a video, compared with that in a still image, can draw support from other frames. Hence, how to aggregate features across different frames is pivotal to VID problem. Most of existing aggregation algorithms are customized for two-stage detectors. However, these detectors are usually computationally expensive due to their two-stage nature. This work proposes a simple yet effective strategy to address the above concerns, which costs marginal overheads with significant gains in accuracy. Concretely, different from traditional two-stage pipeline, we select important regions after the one-stage detection to avoid processing massive low-quality candidates. Besides, we evaluate the relationship between a target frame and reference frames to guide the aggregation. We conduct extensive experiments and ablation studies to verify the efficacy of our design, and reveal its superiority over other state-of-the-art VID approaches in both effectiveness and efficiency. Our YOLOX-based model can achieve promising performance (\emph{e.g.}, 87.5\% AP50 at over 30 FPS on the ImageNet VID dataset on a single 2080Ti GPU), making it attractive for large-scale or real-time applications. The implementation is simple, we have made the demo codes and models available at \url{https://github.com/YuHengsss/YOLOV}.
\end{abstract}

\begin{figure}[!ht]
\centering
\includegraphics[width=1\linewidth]{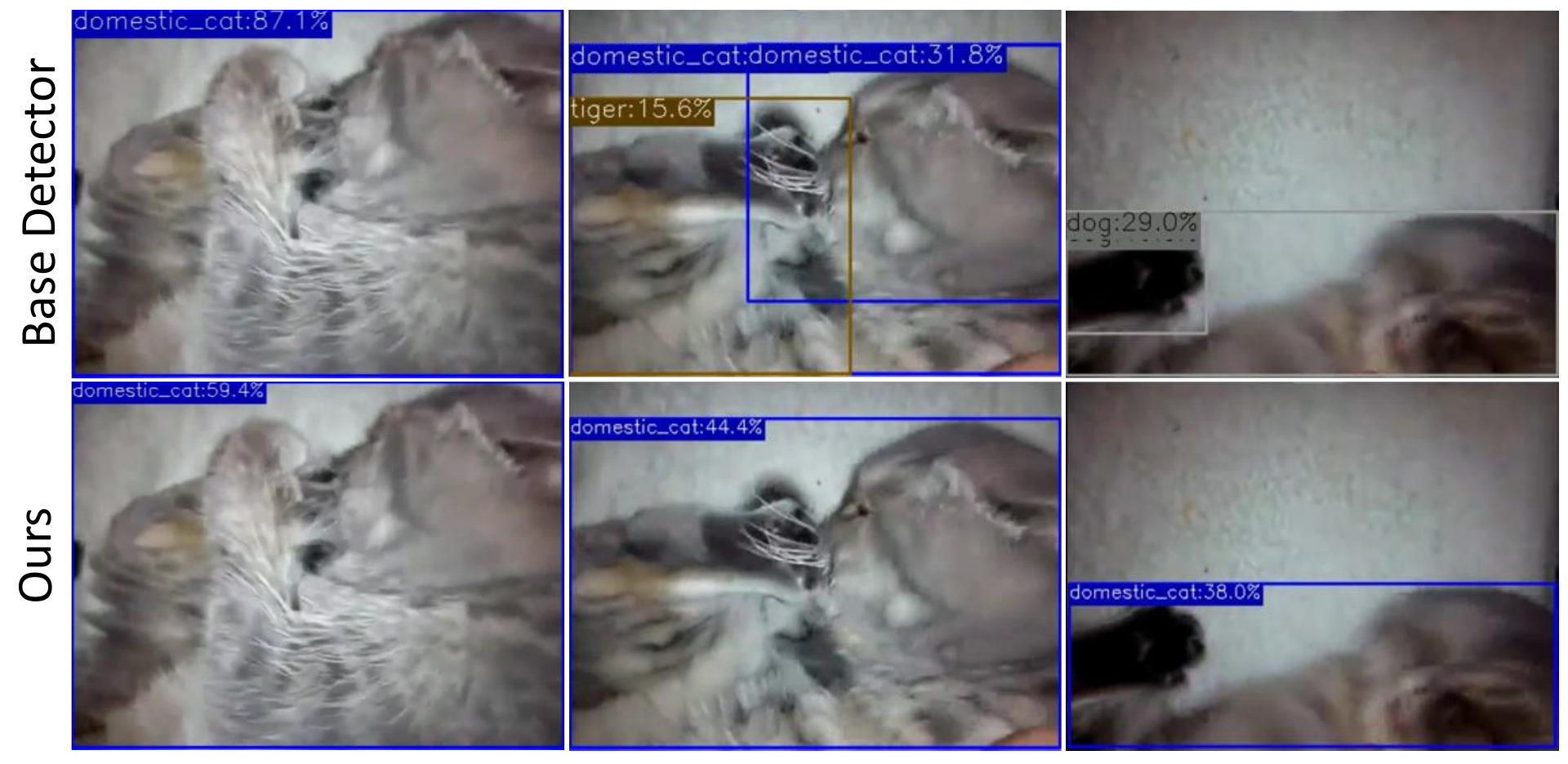}
\caption{These frames suffer from various degradation, like motion blur and occlusion, making the base detector YOLOX fail to accurately detect the objects, while our method could.}
\label{fig:cat_case}

\centering
\includegraphics[width=1\linewidth]{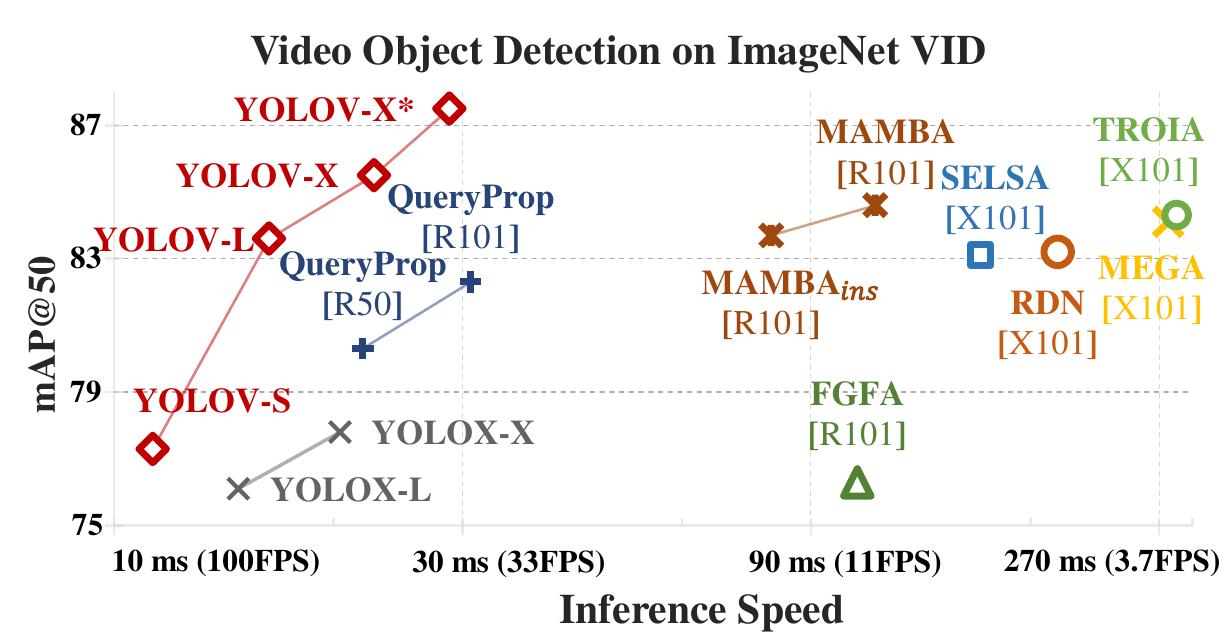}
\caption{Performance comparison in accuracy (AP50) and speed. QueryProp and MBMBA are tested on a TITAN RTX GPU as reported in their papers, while the others are tested on a 2080Ti GPU. $*$ denotes involving post-processing.}
\label{fig:performance}
\end{figure}

\section{Introduction}
Object detection, as a key component in a wide spectrum of vision-based intelligent applications~\cite{dalal2005histograms,felzenszwalb2008discriminatively}, aims to simultaneously locate and classify objects in images. Thanks to the strong ability of Convolutional Neural Networks (CNN)~\cite{krizhevsky2012imagenet}, numerous CNN-based object detection models have been recently proposed, which can be roughly divided into two categories, say one-stage and two-stage object detectors. Specifically, two-stage detectors first select possible object regions (proposals), and then classify these regions. The series of Region-based CNN (R-CNN)~\cite{girshick2014rich,girshick2015fast,ren2015faster} is the pioneer of two-stage object detectors with a variety of follow-ups~\cite{he2016deep,lin2017feature,dai2016r,cai2018cascade,he2017mask,liu2018path}, which remarkably boosts  the accuracy of detection. Given region-level features, these detectors for still images can be easily transferred to more complicated tasks such as segmentation and video object detection. However, due to the two-stage nature, the efficiency is a bottleneck for practical use. While for one-stage object detectors, the localization and classification are jointly and directly produced by the dense prediction from feature maps. The YOLO family~\cite{redmon2016you,redmon2017yolo9000,bochkovskiy2020yolov4} and SSD~\cite{liu2016ssd} are representatives in this group. Without resorting to region proposals as in the aforementioned two-stage approaches, the speed of one-stage detectors is superior and suitable for scenarios with real-time requirements. Though the accuracy of one-stage detectors are typically inferior at the beginning, follow-up designs~\cite{lin2017focal,ge2021yolox,ge2021ota,tian2019fcos} largely narrow the accuracy gap.

Video object detection can be viewed as an advanced version of still image object detection. Intuitively, one can process video sequences via feeding frames one-by-one into still image object detectors. Nevertheless, the temporal information across frames are unexploited, which could be the key to eliminate and reduce the ambiguity in a single image. As shown in Fig.~\ref{fig:cat_case}, degradation such as motion blur, camera defocus, and occlusion often appears in video frames, significantly increasing the difficulty of detection. For instance, via solely looking at the last frame in Fig.~\ref{fig:cat_case}, it is hard or even impossible for human beings to tell where and what the object is. On the other hand, video sequences can provide richer information than single still images. In other words, other frames in the same sequence can possibly support the prediction for a certain frame. Hence, \emph{how to effectively aggregate temporal cues from different frames is crucial to the accuracy}. 

In the literature, there are two main types for frame aggregation, \emph{i.e.}, box-level and feature-level. These two technical routes can boost detection accuracy from different perspectives. Regarding box-level methods, they connect predictions of still object detectors through linking bounding boxes to form tubelets, and then refine the results in the same tubelet. 
The box-level methods can be viewed as post-processing, which are flexible to be applied to both one-stage and two-stage detectors. While for feature-level schemes, the features of a keyframe are enhanced by finding and aggregating similar features from other frames (\emph{a.k.a.}, reference frames). The two-stage manner endows proposals with explicit representation from the backbone feature map extracted by Region Proposal Network (RPN)~\cite{ren2015faster}. Benefiting from this nature, two-stage detectors can be easily migrated to the video object detection problem. Hence, most of video object detectors are built on two-stage detectors. However, because of the aggregation module, these two-stage video object detectors further slow down, and thus hardly meet the need of real-time uses. 
Different from the two-stage methods, proposals are implicitly represented by features of each position in feature maps from a one-stage detector. Though, without explicit representations for objects, these features can still benefit from aggregating temporal information for the VID task. Again, as previously mentioned, the one-stage strategies usually run faster than those two-stage ones. Driven by these considerations, a natural question arises: \emph{Can we make such region-level designs available to one-stage detectors while retaining the fast speed for building a practical (accurate and fast) video object detector}? 

\noindent
\textbf{Contribution.} This paper answers the above question via designing a simple yet effective strategy to aggregate features generated by one-stage detectors\footnote{Please note that our design is general and suitable to many detectors with different backbones.} (\emph{e.g.}, YOLOX~\cite{ge2021yolox} in this work to verify the primary claims). To connect the features of reference frames with those in the keyframe, we propose a feature similarity measurement module to construct an affinity matrix, which is then employed to guide the aggregation. To further mitigate the limitation of cosine similarity, an average pooling operator on reference features is customized. These two operations cost marginal computational resources with significant gains in accuracy. Equipped with the proposed strategies, our model, termed as YOLOV, can achieve a promising accuracy 85.5$\%$ AP50 on the ImageNet VID dataset with 40+ FPS on a single 2080Ti GPU (please see Fig.~\ref{fig:performance} for details) without bells and whistles, which is attractive for practical scenarios. By further introducing post-processing, its accuracy reaches higher up to 87.5$\%$ AP50 over 30 FPS.

\begin{figure*}[t]
\centering
\includegraphics[width=1\linewidth]{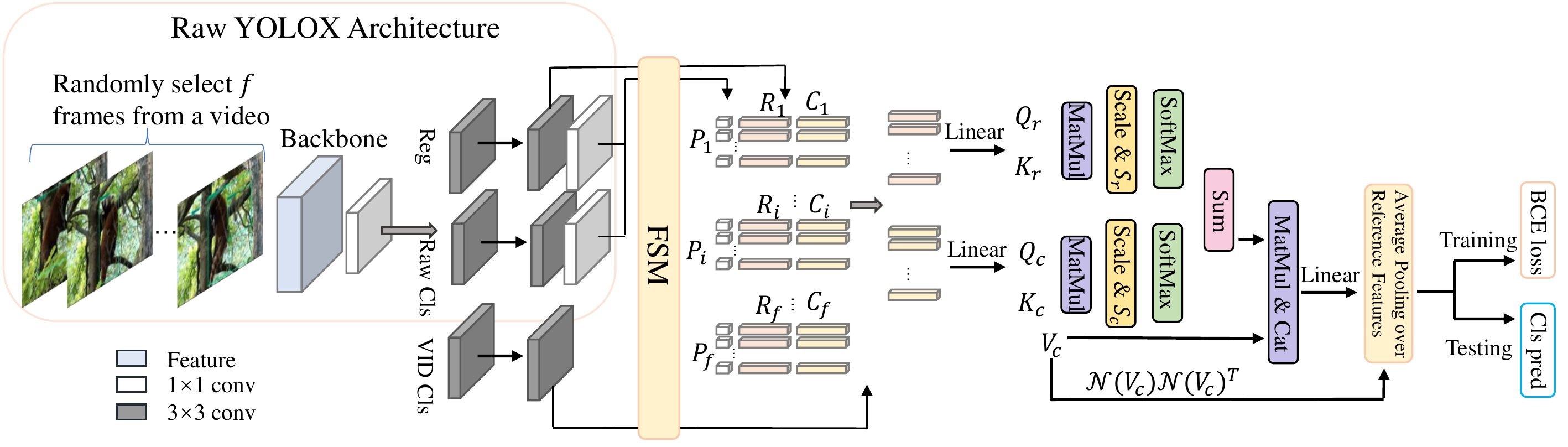}
\caption{Framework of our design. Taking YOLOX as an example base detector, the corresponding model is termed as YOLOV. We randomly sample a number of frames from a video and feed them into the base detector to extract features. According to the predictions of YOLOX, the Feature Selection Module (FSM) picks out top $k$ confident proposals and applies NMS on the selected proposals for further refinement. All the features from FSM are fed into our Feature Aggregation Module (FAM) for final classification. The proposed strategy can be easily applied to other base detectors.}
\label{fig:yolov}
\end{figure*}

\section{Related Work}
This section will briefly review representative approaches in object detection closely related to this work.

\noindent
\textbf{Object Detection in Still Images.} Thanks to the development of hardware, large-scale datasets~\cite{lin2014microsoft,krizhevsky2012imagenet} and sophisticated network structures~\cite{simonyan2014very,he2016deep,xie2017aggregated,wang2020cspnet}, the performance of object detection has continuously improved. Existing object detectors can be mainly divided into two-stage and one-stage schemes. Representative two-stage detectors such as RCNN~\cite{girshick2014rich}, Faster RCNN~\cite{ren2015faster}, R-FCN~\cite{dai2016r}, and Mask RCNN~\cite{he2017mask}. The methods in this group first select candidate regions through RPN and then extract features for the candidates through some feature extraction modules like RoIPooling~\cite{ren2015faster} and RoIAlign~\cite{he2017mask}. Finally, they conduct bounding box regression and classification through an additional detection head. There are also many successful one-stage detectors such as the YOLO series~\cite{redmon2016you,redmon2017yolo9000,bochkovskiy2020yolov4,ge2021yolox}, SSD~\cite{liu2016ssd}, RetinaNet~\cite{lin2017focal}, FCOS~\cite{tian2019fcos}. Different from the two-stage counterparts, the one-stage detectors perform dense prediction on feature maps and directly predict the position and class probability without region proposals.
One-stage detectors are usually faster than two-stage ones, owing to the end-to-end manner. However, they lack explicit region-level semantic features that are widely used for feature aggregation in video object detection. Our work attempts to explore the feasibility of aggregation over position-level features in feature map for one-stage detectors.

\noindent
\textbf{Object Detection in Videos.} Compared to still image object detection, degradation may frequently appear in some video frames. When the keyframe is contaminated, temporal information could be useful for better detection. One branch of existing video object detectors concentrate on tracklet-level post-processing \cite{han2016seq,belhassen2019improving,sabater2020robust}. The methods in this category try to refine the prediction results from the still image detector in consecutive frames by forming the object tubelets. The final classification score of each box is  adjusted according to the entire tubelet. Another branch aims to enhance the features of keyframes, expecting to alleviate degradation via utilizing the features from (selected) reference frames. These approaches can be roughly classified as optical flow-based~\cite{zhu2017flow,zhu2018towards}, attention-based~\cite{wu2019sequence,deng2019relation,chen2020memory,gong2021temporal,sun2021mamba} and tracking-based~\cite{feichtenhofer2017detect,zhang2018integrated} methods. Deep feature flow~\cite{zhu2017deep} first introduces optical flow for image-level feature alignment, and FGFA~\cite{zhu2017flow} adopts optical flow to aggregate features along motion paths. Considering the computational cost of image-level feature aggregation, several attention-based methods have been developed. As a representative, SESLA~\cite{wu2019sequence} proposes a long-range feature aggregation scheme according to the semantic similarity between region-level features. Inspired by the relation module from Relation Networks~\cite{hu2018relation} for still image detection, RDN~\cite{deng2019relation} captures the relationship between objects in both spatial and temporal contexts. Furthermore, MEGA~\cite{chen2020memory} designs a memory enhanced global-local aggregation module for better modeling the relationship between objects. Alternatively, TROIA~\cite{gong2021temporal} utilizes ROI-Align operation for fine-grained feature aggregation, while HVR-Net~\cite{han2020mining} integrates intra-video and inter-video proposal relations for further improvement. Moreover, MBMBA~\cite{sun2021mamba} enlarges the reference feature set by introducing memory bank. QueryProp~\cite{he2022queryprop} notices the high computational cost of video object detectors and tries to speed up the process through a lightweight query propagation module. Besides the attention based methods, D\&T~\cite{feichtenhofer2017detect} tries solving video object detection in a tracking manner by constructing correlation maps of different frame features. Although these approaches boost the precision of detection, they are mostly based on two-stage detectors and hence suffer from the relatively slow inference speed.

\section{Methodology}

Considering the characteristics of videos (various degradation \emph{vs.} rich temporal information), instead of individually processing frames, how to seek supportive information from other frames for a target one (keyframe) plays a key role in boosting the accuracy of video object detection. Recent attempts~\cite{deng2019relation,chen2020memory,wu2019sequence,he2022queryprop} with noticeable improvement in accuracy corroborate the importance of temporal aggregation to the problem. However, most of the existing methods are two-stage based techniques. 
As previously discussed, their main drawback is relatively slow speed, compared to one-stage bases. To mitigate this limitation, we put the region/feature selection after the prediction head of one-stage detectors. In this section, we choose the YOLOX as base to present our main claims. Our proposed framework is schematically depicted in Fig.~\ref{fig:yolov}. Let us recall the traditional two-stage pipeline: 1) massive candidate regions are first ``selected" as proposals; and 2) determine if each proposal is an object or not and which class it belongs to. The computational bottleneck mainly comes from dealing with substantial low-confidence region candidates. As can be seen from Fig.~\ref{fig:yolov}, our pipeline also contains two stages. Differently, \emph{its first stage is prediction (having a large number of regions with low confidences discarded), while the second stage can be viewed as region-level refinement (taking advantage of other frames by aggregation).} By this principle, our design can simultaneously benefit from the efficiency of one-stage detectors and the accuracy gained from temporal aggregation. It is worth to emphasize that \emph{such a small difference in design leads to a huge difference in performance.} The proposed strategy can be generalized to many base detectors such as FCOS~\cite{tian2019fcos} and PPYOLOE~\cite{xu2022pp}.

\subsection{Our Design}
From a possible perspective of human beings, the recognition procedure would first link \emph{related} instances in temporal and identify which class they belong to until sufficiently \emph{confident} cues are collected. Then, the results can be broadcasted to less confident cases. The mechanism of multi-head attention, as a key part of Transformers~\cite{vaswani2017attention}, seems to fit the situation well, which enhances the capability of long-range modeling. Given a sequence $Z$, the query, key, and value matrices are packed as $Q, K$ and $V$, respectively. The self-attention can be calculated via:
\begin{equation}
 \operatorname{SA}(Z) =  \operatorname{softmax}\left(A\right)V\quad \text{with} \quad A =QK^{T}/{\sqrt{d}},
\end{equation}
where $d$ is the dimension of each feature in Q (also K).
Putting $m$ self-attentions in parallel yields the multi-head attention by simply concatenating them together as follows:
\begin{equation}
\operatorname{MSA}(Z)=\operatorname{concat}\big(\operatorname{SA}_1(Z),\operatorname{SA}_2(Z),...,\operatorname{SA}_m(Z)\big).
\end{equation}
Modern two-stage based video object detectors typically obtain candidate regions for feature aggregation by RPN~\cite{ren2015faster}. As a representative, RelationNet~\cite{hu2018relation} first introduces the above multi-head attention to still object detection task by viewing a sequence of proposals as input. The ROI-Pooling or ROI-Align operations are applied to these proposals to extract region-level features. However, one-stage detectors make dense prediction directly from feature maps. Simply transferring the region-level feature aggregation to the whole feature maps of one-stage detectors will result in intensive computational cost. To address this issue, we propose an effective strategy for selecting features that suitable for multi-head attentions.

\subsubsection{FSM: Feature Selection Module}
As most of the predictions are of low confidence, the detection head of one-stage detectors is a natural and rational option to select (high-quality) candidates from the feature maps. Following the process of RPN, we first pick out top $k$ (\emph{e.g.}, 750) predictions according to the confidence scores. Then, a fixed quantity $a$ of predictions (\emph{e.g.}, $a=30$) are chosen after the Non-Maximum Suppression (NMS) for reducing the redundancy. The features of these predictions will be collected for further refinement. 
In practice, we found that directly aggregating the collected features in the classification branch and backpropagating the classification loss of the aggregated features will result in unstable training. Since the weight of the feature aggregation module is randomly initialized, finetuning all the networks from the beginning will contaminate the pre-trained weights. To address the above concerns, we fix the weights in the base detector except for the linear projection layers in detection head. We further insert two $3 \times 3$ convolutional (Conv) layers into the model neck as a new branch, called video object classification branch, which generates features for aggregation. Then, we feed the collected features from the video and regression branches into our feature aggregation module. 

\subsubsection{FAM: Feature Aggregation Module}

Now we come to the step of connecting related RoIs.
Let $\mathcal{F} = \left\{{C}_{1}, {C}_{2}, ..., {C}_{f};{R}_{1}, {R}_{2}, ..., {R}_{f}\right\}$ denote the feature set selected by FSM where $C_i\in\mathbb{R}^{d_q\times a} = \left[\mathbf{c}_{i}^{1}, \mathbf{c}_{i}^{2}, ..., \mathbf{c}_{i}^{a}\right]$ and $R_i\in\mathbb{R}^{d_q\times a} = \left[\mathbf{r}_{i}^{1}, \mathbf{r}_{i}^{2}, ..., \mathbf{r}_{i}^{a}\right]$ denotes the features of the $i$-th frame in $\mathcal{F}$ from the video classification and regression branches, respectively. $d_q$ refers to the feature dimension of each RoI and $f$ refers to the number of related frames. The generalized cosine similarity is arguably the most widely used metric to compute the similarity between features or the attention weight~\cite{wu2019sequence,shvets2019leveraging,deng2019relation}. Simply referring to the cosine similarity will find features most similar to the target. However, when the target suffers from some degradation, the selected reference proposals using cosine similarity are very likely to have the same problem. We name this phenomenon \emph{the homogeneity issue}. 

First, let's recall the original \textbf{QK manner} in self-attention for our method. Similarly to~\cite{vaswani2017attention}, the query, key, and value matrices are constructed and fed into the multi-head attention. 
For instance, $Q_c$ and $Q_r$ are respectively formed by stacking the features from the classification branch and the regression branch for all proposals in all related frames (\emph{i.e.}, $Q_c\in\mathbb{R}^{fa\times d}=\operatorname{LP}([{C}_{1}, {C}_{2}, ..., {C}_{f}]^T)$ and $Q_r\in\mathbb{R}^{fa\times d}=\operatorname{LP}([{R}_{1}, {R}_{2}, ..., {R}_{f}]^T)$, where $\operatorname{LP}(\cdot)$ is the linear projection operator which projects the features of $d_q$-dim to $d$-dim), while the others are done similarly.
By the scaled dot-product in attention, we obtain the corresponding $A_c = Q_c K_c^{T} /\sqrt{d} $ and $A_r = Q_r K_r^{T} /\sqrt{d}$.  

To overcome the homogeneity issue, we further take predicted confidences from the raw detector into consideration, denoted as $\mathbf{P} \in \mathbb{R}^{2 \times fa} = \left\{{P}_{1}, {P}_{2}, ..., {P}_{f}\right\}$ with each $P_i$ contains two scores, \emph{i.e.}, the classification score and IoU score from the raw classification and regression heads. 
To fit these scores into the attention weights, we build two matrices, say $S_r\in\mathbb{R}^{fa\times fa}$ and $S_c\in\mathbb{R}^{fa\times fa}$, through respectively repeating each of two rows of $\mathbf{P}$ $fa$ times. As a consequence, the self-attentions for the classification and regression branches turn out to be:
\begin{equation}
\begin{array}{c}
\operatorname{SA_c}(\mathcal{C})=\operatorname{softmax}\left({S_c \circ A_c }\right) V_c,\\
\operatorname{SA_r}(\mathcal{R})=\operatorname{softmax}\left({S_r \circ A_r }\right) V_c,
\label{eq:msa}
\end{array}
\end{equation}
where $\circ$ represents the Hadamard product. 
By this operation, the self-attention considers not only the similarity between the query and key items, but also aware of the quality of the keys. Note that, since the main purpose is to refine the classification, $\operatorname{SA_c}(\mathcal{C})$ and $\operatorname{SA_r}(\mathcal{R})$ share the same value matrix $V_c$.
Our experiments demonstrate that replacing the original QK manner with Eq. \eqref{eq:msa} (called \textbf{affinity manner}) in multi-head attention can significantly boost the performance of video object detector. Besides, we concatenate $V_c$ with the outputs of Eq. \eqref{eq:msa} for better preserving initial representations via:
 \begin{equation}
\begin{array}{c}
\operatorname{SA(\mathcal{F})}= \operatorname{concat}((\operatorname{SA_c}(\mathcal{C})+\operatorname{SA_r}(\mathcal{R}),V_c).
\end{array}
\end{equation} 
The positional information is not embedded, because the locations in a long temporal range would not be helpful as claimed in~\cite{chen2020memory}.

\begin{figure}[t]
\centering
\includegraphics[width=1\linewidth]{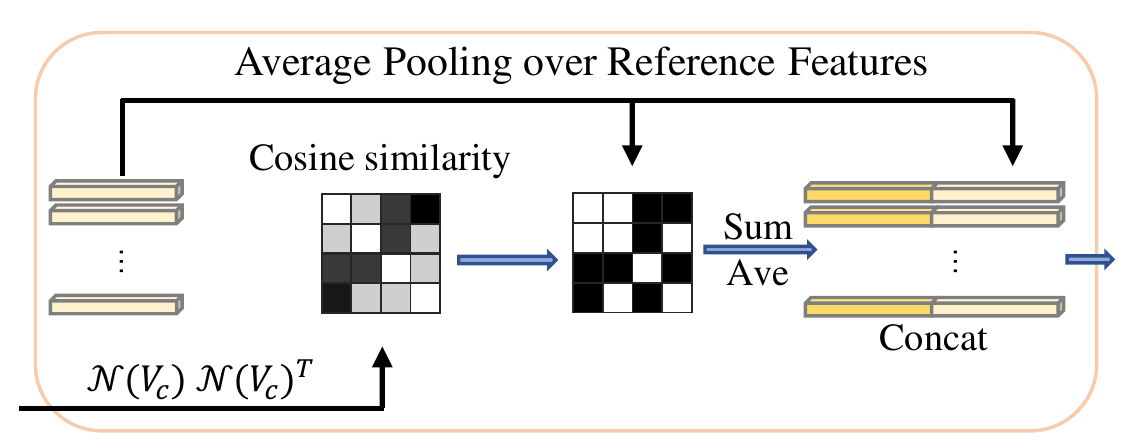}
\caption{Our average pooling over reference features.}
\label{fig:ave pooling fig}
\end{figure}

Moreover, considering the characteristic of softmax, a small part of references may hold a large portion of weights. In other words, it often overlooks the references with low weights, which limits the diversity of reference features for possible follow-up use. To avoid such risks, we introduce an \textbf{average pooling over reference features (A.P.)}. Concretely, we select all of the references with similarity scores above a threshold $\tau$ and apply the average pooling to these survivals. Note that the similarity in this work is computed via $\mathcal{N}(V_c) \mathcal{N}(V_c)^{T}$. The operator $\mathcal{N}(\cdot)$ means the layer normalization, which guarantees the values to be in a certain range and thus eliminates the influence of scale difference.
The average-pooled features and the key features are then fed into one linear projection layer for final classification. The procedure is illustrated in Fig.~\ref{fig:ave pooling fig}. 

\section{Experimental Validation}

\subsection{Implementation Details}

Similar to previous work, we also initialize our base detector from COCO pre-trained weights provided by YOLOX. Following~\cite{zhu2017flow,wu2019sequence,gong2021temporal}, we combine the videos in the ImageNet VID and those in the ImageNet DET with the same classes as our training data. Specifically, the ImageNet VID ~\cite{russakovsky2015imagenet} contains 3,862 videos for training and 555 videos for validation. There are 30 categories in the VID dataset, \emph{i.e.}, a subset of the 200 basic-level categories of ImageNet DET~\cite{russakovsky2015imagenet}.  Considering the redundancy of video frames, we randomly sample 1/10 frames in the VID training set instead of using all of them. The base detectors are trained for 7 epochs by SGD with batch size of 16 on 2 GPUs. As for the learning rate, we adopt the cosine learning rate schedule used in YOLOX with one epoch for warming up and disable the strong data augmentation for the last 2 epochs. When integrating the feature aggregation module into the base detectors, we fine-tune them for 150K iterations with batch size of 16 on a single 2080Ti GPU. In addition, we use warm-up for the first 15K iterations and cosine learning rate schedule for the rest iterations. Only the linear projection layers in YOLOX prediction head, the newly added video object classification branch and the multi-head attention are fine-tuned. For the training of feature aggregation module, the number of frames $f$ is set to 16, and the threshold of NMS is set to $0.75$ for rough feature selection. While for producing final detection boxes, we alternatively set the threshold of NMS to $0.5$ for retaining more confident candidates. In the training phase, the images are randomly resized from $352\times 352$ to $672\times 672$ with 32 steps. In the testing phase, the images are uniformly resized to $576\times576$. The AP50 and inference speed are two metrics to reflect the performance in terms of accuracy and efficiency, respectively. Regarding the inference speed, we test all of the models with FP16-precision on a 2080Ti GPU unless otherwise stated.

\subsection{Ablation Study}
\subsubsection{On the reference frame sampling strategy.} Investigating frame sampling strategies to balance accuracy and efficiency is crucial for video object detectors. Several global and local sampling schemes have been discussed in previous two-stage based methods~\cite{wu2019sequence,gong2021temporal,chen2020memory}. As for the global sampling schemes, $f_g$ frames are randomly selected from the whole video. With respect to the local sampling, $f_l$ consecutive frames around the keyframe are employed. To illustrate the effect of different sampling strategies, we vary the numbers of reference frames in both global and local sampling. The numerical results are reported in Table~\ref{table:effectiveness of local and global number}. The performance of using only 3 global reference frames has already outperformed that of using 39 local reference frames, which corroborates the evidence given in ~\cite{wu2019sequence,gong2021temporal}. As a trade-off, we adopt the global sampling strategy with $f_g= 31$ by default for the remaining experiments according to Table~\ref{table:effectiveness of local and global number}. 

\begin{figure*}[t]
\centering
\includegraphics[width=1\linewidth]{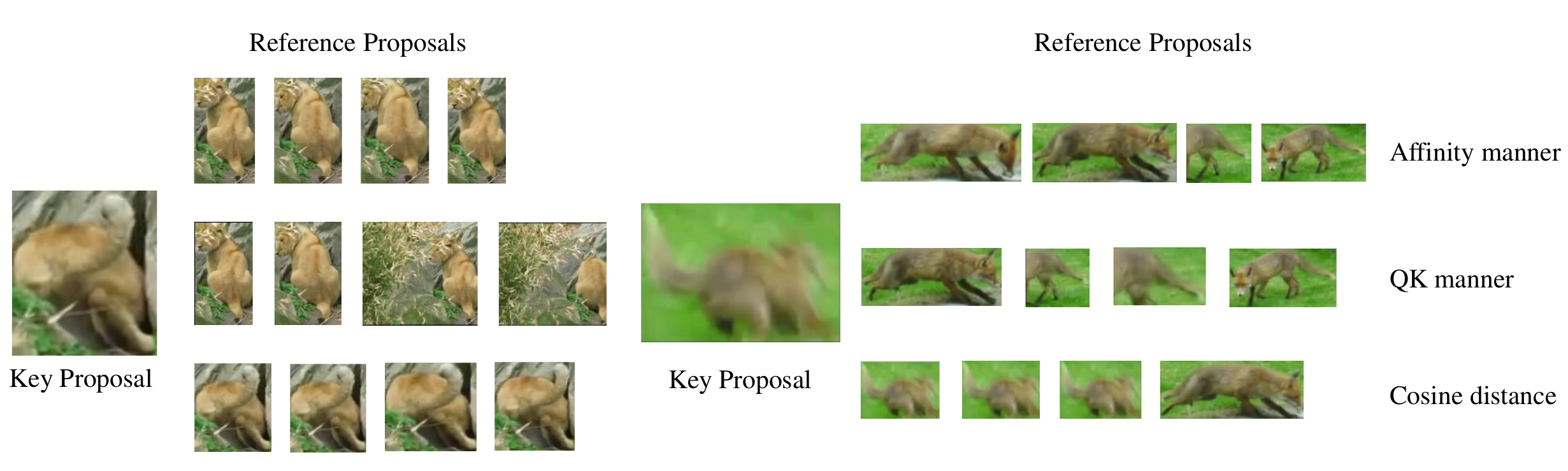}
\caption{Visual comparison between reference proposals selected by three different methods for given key proposals. We display four reference proposals that contribute most in aggregation.}
\vspace{1mm}
\label{fig:conf compare}
\end{figure*}

\setlength{\tabcolsep}{4pt}

\begin{table}[t]
\begin{center}
\begin{tabular}{c|cccccc}
\hline\noalign{\smallskip}
$f_g$ & 3 & 7 & 15 & 23 & \textbf{31} & 39\\
\noalign{\smallskip}
\hline
\noalign{\smallskip}
AP50 ($\%$) & 74.6 & 76.1 & 77.0 & 77.3 & \textbf{77.3} & 77.3 \\
\hline
\hline\noalign{\smallskip}
$f_l$ & 3 & 7 & 15 & 23 & 31 & 39\\
\noalign{\smallskip}
\hline
\noalign{\smallskip}
AP50 ($\%$) & 70.4 & 71.2 & 72.2 & 72.8 & 73.2 & 73.6 \\
\hline
\end{tabular}
\end{center}
\caption{Effect of the numbers of global $f_g$ and local $f_l$ reference frames.}
\label{table:effectiveness of local and global number}
\end{table}

\setlength{\tabcolsep}{4pt}
\begin{table}
\begin{center}
\begin{tabular}{c|cccccc}
\hline\noalign{\smallskip}
$a$ & 10 & 20 & \textbf{30} & 50 & 75 & 100\\
\noalign{\smallskip}
\hline
\noalign{\smallskip}
AP50 ($\%$) & 76.1 & 77.0 & \textbf{77.3} & 77.3 & 77.4 & 77.4 \\
\hline
\noalign{\smallskip}
Time (ms) & 10.4 & 10.7 & \textbf{11.3} & 13.8 & 19.5 & 29.3 \\
\hline
\end{tabular}
\end{center}
\caption{Effect of the number of proposals $a$ in FSM.}
\label{table:effectiveness of a in fsm}
\end{table}

\subsubsection{On the number of proposals in a single frame.}
In this experiment, we adjust the quantity of most confident proposals remained for each frame $a$ from 10 to 100 in the FSM, to see its influence on the performance. As shown in Table~\ref{table:effectiveness of a in fsm}, as $a$ increases, the accuracy continuously improves and saturates until reaching 75. As the complexity of self-attention is $O\left(n^{2}\right)$ with respect to the amount of input proposals, involving too many proposals per frame will dramatically increase the time cost. As a tradeoff, we adopt $a=30$ that is much smaller than the optimal setting with 75 proposals used in the two-stage based method RDN~\cite{deng2019relation}.

\subsubsection{On the threshold in average pooing of reference features.}
Here, we test the effect of different thresholds for average pooling over reference features. Table~\ref{table:effectiveness of thresh in average} lists the numerical results. When all the features are involved in the average pooling, \emph{i.e.} $\tau=0$, the AP50 is merely $73.1\%$. Raising to higher threshold results in better performance. When $\tau$ falls in $[0.75, 0.85]$, the accuracy keeps steady high to $77.3\%$. But, when $\tau=1$, the average pooling is equivalent to only duplicating $\operatorname{SA(\mathcal{F})}$, the accuracy of which drops to $76.9\%$. Dynamically determining the threshold for different cases is desired and is left as our future work. For the rest of experiments, we adopt $\tau=0.75$ as the default threshold. 
\setlength{\tabcolsep}{4pt}
\begin{table}[t]
\begin{center}
\begin{tabular}{c|ccccccc}
\hline\noalign{\smallskip}
 $\tau$ & 0 & 0.2 & 0.5 & 0.65 & \textbf{0.75} & 0.85 & 1\\
\noalign{\smallskip}
\hline
\noalign{\smallskip}
AP50 ($\%$) & 73.1 & 73.9 & 76.1 & 77.1 & \textbf{77.3} & 77.3 & 76.9 \\
\hline
\end{tabular}
\end{center}
\caption{Influence of the threshold $\tau$ in average pooling over reference features.}
\label{table:effectiveness of thresh in average}
\end{table}

\setlength{\tabcolsep}{4pt}
\begin{table}[t]
\begin{center}
\begin{tabular}{c|cccc}
\hline\noalign{\smallskip}
Methods&A.M. &A.P.&Time (ms)&AP50 ($\%$)\\
\noalign{\smallskip}
\hline
\noalign{\smallskip}
YOLOX-S & - & - & 9.38 &$69.5$\\
YOLOV-S & \checkmark & - & 10.95 & $76.9_{\uparrow 7.4 }$ \\
{YOLOV-S} & \checkmark& \checkmark & {11.30} & {$77.3_{\uparrow 7.8 }$}\\
\hline
\end{tabular}
\end{center}
\caption{Effectiveness of our affinity manner (A.M.) and average pooling over reference features (A.P.). }
\label{table:effectiveness of ours}
\end{table}
\setlength{\tabcolsep}{1.4pt}

\setlength{\tabcolsep}{4pt}
\begin{table}[t]
\begin{center}
\begin{tabular}{c|cccc}
\hline\noalign{\smallskip}
Model&Params&GFLOPs&Time (ms)&AP50 ($\%$)\\
\noalign{\smallskip}
\hline
\noalign{\smallskip}

YOLOX-S & 8.95M & 21.63 & 9.4 &$69.5$\\
YOLOV-S & 10.28M & 26.18 & 11.3 & $77.3_{\uparrow 7.8 }$\\
YOLOX-L & 54.17M & 125.90 & 14.8 &  $76.1$\\
YOLOV-L & 59.45M & 143.10 & 16.4 &  $83.6_{\uparrow 7.5}$ \\
YOLOX-X & 99.02M & 228.15 & 20.4 & $77.8$\\
YOLOV-X & 107.26M & 254.72 & 22.7 & $85.0_{\uparrow 7.2 }$\\
YOLOV-X$^\dagger$ & 107.26M & 254.72 & 22.7 & $85.5_{\uparrow 7.7 }$\\

\hline
\end{tabular}
\end{center}
\caption{Effectiveness of our strategy compared to bases. $\dagger$ indicates using strong augmentation.}
\label{table:effectiveness of diff model}
\end{table}
\setlength{\tabcolsep}{1.4pt}

\setlength{\tabcolsep}{4pt}
\begin{table}[t]
\begin{center}
\begin{tabular}{l|ccccc}
\hline\noalign{\smallskip}
Method & Backbone  & AP50 ($\%$) & Time (ms)  \\
\hline
\noalign{\smallskip}
FGFA & R101   & 76.3 & 104.2  \\
SELSA & X101   & 83.1 & 153.8  \\ 
RDN & X101   & 83.2 & -  \\
MEGA & R101   & 82.9 & 230.4  \\
MEGA & X101   & 84.1 & -  \\
TROIA & X101   & 84.3 & 285.7  \\
MAMBA & R101   & 84.6 & 110.3$(T)$  \\
HVR & X101   & 84.8 & -  \\
TransVOD & R101   & 81.9 & -  \\
QueryProp & R50  & 80.3 & 21.9$(T)$  \\
QueryProp & R101  & 82.3 & 30.8$(T)$  \\
\hline
\noalign{\smallskip}
\textbf{YOLOV-R50} & R50    & \textbf{81.4} & \textbf{17.5}  \\
\textbf{YOLOV-S} & MCSP    & \textbf{77.3} & \textbf{11.3}  \\
\textbf{YOLOV-L} & MCSP    & \textbf{83.6} & \textbf{16.3}  \\
\textbf{YOLOV-X} & MCSP    & \textbf{85.0} & \textbf{22.7}  \\
\textbf{YOLOV-X$^\dagger$} & MCSP   & \textbf{85.5} & \textbf{22.7}  \\
\hline
\hline
\noalign{\smallskip}
FGFA & R101   & 78.4 & - \\
SELSA & X101   & 83.7 & - \\
RDN & X101   & 84.7 & - \\
MEGA & X101   & 85.4 & - \\
HVR & X101   & 85.5 & -\\
\hline
\noalign{\smallskip}
\textbf{YOLOV-S} & MCSP    & \textbf{80.1} & \textbf{11.3 + 6.9} \\
\textbf{YOLOV-L}& MCSP    & \textbf{86.2} & \textbf{16.3 + 6.9}\\
\textbf{YOLOV-X}& MCSP    & \textbf{87.2} & \textbf{22.7 + 6.1}\\
\textbf{YOLOV-X$^\dagger$} & MCSP   & \textbf{87.5} & \textbf{22.7 + 6.1}\\
\hline
\end{tabular}
\end{center}
\caption{Performance comparison in accuracy and speed. $\dagger$ indicates with strong augmentation, $T$ means the inference time is tested on a TITAN RTX GPU as reported in corresponding papers. MCSP stands for the Modified CSP v5 backbone adopted in YOLOX. The lower part involves post-processing while the upper does not.}
\label{table:compare w post}
\end{table}

\subsubsection{On the effectiveness of FAM.} To validate the effectiveness of the affinity manner (A.M.) and the average pooling over reference features (A.P.), we evaluate the performance with and without these modules. The results in Table~\ref{table:effectiveness of ours} reveal that these designs can both help the feature aggregation to catch better semantic representations from one-stage detectors. Compared to YOLOX-S ($69.5\%$ AP50), the YOLOV-S only armed with A.M. gains $7.4\%$ in accuracy. While equipping YOLOV-S with both A.M. and A.P. (complete YOLOV-S), the performance reaches $77.3\%$ in AP50 with only about 2ms time cost in comparison with YOLOV-S. Furthermore, we also plug our design into the YOLOX-L and YOLOX-X. Tabel~\ref{table:effectiveness of diff model} shows the detailed comparison. The $\dagger$ indicates using strong augmentation (like MixUp~\cite{zhang2018mixup} and Mosaic~\cite{bochkovskiy2020yolov4}) when fine-tuning our version. Our YOLOV consistently outperform their corresponding baselines over $7\%$ in AP50. 

Moreover, we provide two cases to intuitively exhibit the superiority of our FAM. They are the lion case with rare pose and the fox case with motion blur as shown in Fig~\ref{fig:conf compare}. Without loss of generality, the top $4$ reference proposals are listed for different feature selection modes, including the cosine similarity, QK manner in multi-head attention, and our affinity manner. As previously analyzed, the cosine manner selects proposals most similar to the key proposal but suffering the same degradation problem as the key proposal. Though the QK manner alleviates this problem, it is obviously inferior to the affinity manner. By introducing the confidence scores as guidance, our method selects better proposals and further boosts the detection accuracy.

\setlength{\tabcolsep}{4pt}
\begin{table}[t]
\begin{center}
\begin{tabular}{l|ccl}
\hline\noalign{\smallskip}
Model&Params&GFLOPs&AP50 ($\%$)\\
\noalign{\smallskip}
\hline
\noalign{\smallskip}
PPYOLOE-S & 6.71M & 12.40 &$69.5$\\
PPYOLOE-S+Ours & 8.04M & 16.73 & $74.9_{\uparrow 5.6}$\\
PPYOLOE-L & 50.35M & 89.19  &  $76.9$\\
PPYOLOE-L+Ours & 55.63M & 105.49 &  $82.0_{\uparrow 5.1}$ \\
FCOS & 31.00M & 102.84& $67.0$\\
FCOS+Ours & 36.28M & 120.04 & $73.1_{\uparrow 6.1 }$\\
\hline
\end{tabular}
\end{center}
\caption{Effectiveness of our strategy on other base detectors.}
\label{table:effectiveness of diff detector}
\end{table}
\setlength{\tabcolsep}{1.4pt}

\subsection{Comparison with State-of-the-art Methods}
Table~\ref{table:compare w post} summarizes the detailed performance of related methods including FGFA~\cite{zhu2017flow}, SELSA~\cite{wu2019sequence}, RDN~\cite{deng2019relation}, MEGA~\cite{chen2020memory}, TROIA~\cite{gong2021temporal}, MAMBA~\cite{sun2021mamba}, HVR~\cite{han2020mining}, TransVOD~\cite{he2021end} and QueryProp~\cite{he2022queryprop}. Our method can achieve $85.5\%$ AP50 with 21.1 ms per frame. With the help of a recent post-processing method REPP~\cite{sabater2020robust}, it reaches $87.5\%$ AP50 spending extra 6 ms. In terms of inference efficiency, our method is much faster than the other methods. To show the improvement is not purely from changing to stronger backbone network, we also report the speed and accuracy using ResNet50.

To be specific, in the upper part of Table~\ref{table:compare w post}, we report the performance without any post-processing methods. YOLOV can remarkably lead in both detection accuracy and inference efficiency. For fair comparison, all the models listed in Table~\ref{table:compare w post} are tested with the same hardware environment except for MAMBA and QueryProp\footnote{The codes and models of the methods with marks `T' (the reported results are from their papers). '-' denotes the model is not released and the speed is not reported in orginal paper.}.  The lower part of Table~\ref{table:compare w post} reports the results of our YOLOV and the other SOTA models with post-processing. The time cost of post-processing is tested on an i7-8700K CPU.

\subsection{Application to Other Base Detectors}
In order to validate the generalization ability of the proposed strategy, we also try it on other widely-used one-stage detectors including PPYOLOE~\cite{xu2022pp} and FCOS~\cite{tian2019fcos}. Specifically for PPYOLOE, it has different channel numbers at different FPN levels. To achieve the multi-scale feature aggregation, we reduce the channel numbers of the detection head at different scales to be the smallest of all levels in the pre-trained model. While for FCOS, the backbone is ResNet-50~\cite{he2016deep}. The FPN consists of five levels in the original architecture for dealing with images with large image sizes (\emph{e.g.}, $1333\times 800$). To match the situation of the ImageNet VID, we maintain three FPN levels with the largest downsampling rate of 32. For the training procedure and other hyper-parameters settings, we simply keep them the same as those in the YOLOX. Table~\ref{table:effectiveness of diff detector} shows that our strategy can consistently improve different base detectors by over $5\%$ in terms of AP50. It is worth noting that searching for more suitable hyper-parameters for different base detectors could obtain better performance.\\


\section{Conclusion}
\vspace{2mm}
In this paper, we developed a practical video object detector both efficiently and effectively. To improve the detection accuracy, a feature aggregation module was designed to aggregate temporal information. For saving computational resources, different from existing two-stage detectors, we proposed to put the region selection after the one-stage prediction. This subtle change makes our detectors significantly more efficient. 
Experiments and ablation studies have been conducted to verify the effectiveness of our strategy, and it surpasses over previous arts. The core idea is simple and general, which can potentially inspire further research works and broaden the applicable scenarios related to video object detection.

\section*{Acknowledgments}
This work was supported by the National Natural Science Foundation of China under Grant no. 62072327.

\bibliography{ShiYuheng}

\begin{thebibliography}{46}
\providecommand{\natexlab}[1]{#1}

\bibitem[{Belhassen et~al.(2019)Belhassen, Zhang, Fresse, and
  Bourennane}]{belhassen2019improving}
Belhassen, H.; Zhang, H.; Fresse, V.; and Bourennane, E.-B. 2019.
\newblock Improving Video Object Detection by Seq-Bbox Matching.
\newblock In \emph{VISIGRAPP (5: VISAPP)}, 226--233.

\bibitem[{Bochkovskiy, Wang, and Liao(2020)}]{bochkovskiy2020yolov4}
Bochkovskiy, A.; Wang, C.-Y.; and Liao, H.-Y.~M. 2020.
\newblock Yolov4: Optimal speed and accuracy of object detection.
\newblock \emph{arXiv preprint arXiv:2004.10934}.

\bibitem[{Cai and Vasconcelos(2018)}]{cai2018cascade}
Cai, Z.; and Vasconcelos, N. 2018.
\newblock Cascade R-CNN: Delving into high quality object detection.
\newblock In \emph{CVPR}, 6154--6162.

\bibitem[{Chen et~al.(2020)Chen, Cao, Hu, and Wang}]{chen2020memory}
Chen, Y.; Cao, Y.; Hu, H.; and Wang, L. 2020.
\newblock Memory enhanced global-local aggregation for video object detection.
\newblock In \emph{CVPR}, 10337--10346.

\bibitem[{Dai et~al.(2016)Dai, Li, He, and Sun}]{dai2016r}
Dai, J.; Li, Y.; He, K.; and Sun, J. 2016.
\newblock R-fcn: Object detection via region-based fully convolutional
  networks.
\newblock In \emph{NIPS}, 379--387.

\bibitem[{Dalal and Triggs(2005)}]{dalal2005histograms}
Dalal, N.; and Triggs, B. 2005.
\newblock Histograms of oriented gradients for human detection.
\newblock In \emph{CVPR}, 886--893.

\bibitem[{Deng et~al.(2019)Deng, Pan, Yao, Zhou, Li, and
  Mei}]{deng2019relation}
Deng, J.; Pan, Y.; Yao, T.; Zhou, W.; Li, H.; and Mei, T. 2019.
\newblock Relation distillation networks for video object detection.
\newblock In \emph{ICCV}, 7023--7032.

\bibitem[{Feichtenhofer, Pinz, and Zisserman(2017)}]{feichtenhofer2017detect}
Feichtenhofer, C.; Pinz, A.; and Zisserman, A. 2017.
\newblock Detect to track and track to detect.
\newblock In \emph{ICCV}, 3038--3046.

\bibitem[{Felzenszwalb, McAllester, and
  Ramanan(2008)}]{felzenszwalb2008discriminatively}
Felzenszwalb, P.; McAllester, D.; and Ramanan, D. 2008.
\newblock A discriminatively trained, multiscale, deformable part model.
\newblock In \emph{CVPR}, 1--8.

\bibitem[{Ge et~al.(2021{\natexlab{a}})Ge, Liu, Li, Yoshie, and
  Sun}]{ge2021ota}
Ge, Z.; Liu, S.; Li, Z.; Yoshie, O.; and Sun, J. 2021{\natexlab{a}}.
\newblock Ota: Optimal transport assignment for object detection.
\newblock In \emph{CVPR}, 303--312.

\bibitem[{Ge et~al.(2021{\natexlab{b}})Ge, Liu, Wang, Li, and
  Sun}]{ge2021yolox}
Ge, Z.; Liu, S.; Wang, F.; Li, Z.; and Sun, J. 2021{\natexlab{b}}.
\newblock Yolox: Exceeding yolo series in 2021.
\newblock \emph{arXiv preprint arXiv:2107.08430}.

\bibitem[{Girshick(2015)}]{girshick2015fast}
Girshick, R. 2015.
\newblock Fast R-CNN.
\newblock In \emph{ICCV}, 1440--1448.

\bibitem[{Girshick et~al.(2014)Girshick, Donahue, Darrell, and
  Malik}]{girshick2014rich}
Girshick, R.; Donahue, J.; Darrell, T.; and Malik, J. 2014.
\newblock Rich feature hierarchies for accurate object detection and semantic
  segmentation.
\newblock In \emph{CVPR}, 580--587.

\bibitem[{Gong et~al.(2021)Gong, Chen, Wang, Chu, Zhu, Lin, Yu, and
  Feng}]{gong2021temporal}
Gong, T.; Chen, K.; Wang, X.; Chu, Q.; Zhu, F.; Lin, D.; Yu, N.; and Feng, H.
  2021.
\newblock Temporal ROI align for video object recognition.
\newblock In \emph{AAAI}, 1442--1450.

\bibitem[{Han et~al.(2020)Han, Wang, Chang, and Qiao}]{han2020mining}
Han, M.; Wang, Y.; Chang, X.; and Qiao, Y. 2020.
\newblock Mining inter-video proposal relations for video object detection.
\newblock In \emph{ECCV}, 431--446.

\bibitem[{Han et~al.(2016)Han, Khorrami, Paine, Ramachandran, Babaeizadeh, Shi,
  Li, Yan, and Huang}]{han2016seq}
Han, W.; Khorrami, P.; Paine, T.~L.; Ramachandran, P.; Babaeizadeh, M.; Shi,
  H.; Li, J.; Yan, S.; and Huang, T.~S. 2016.
\newblock Seq-NMS for video object detection.
\newblock \emph{arXiv preprint arXiv:1602.08465}.

\bibitem[{He et~al.(2022)He, Gao, Jia, Zhao, and Huang}]{he2022queryprop}
He, F.; Gao, N.; Jia, J.; Zhao, X.; and Huang, K. 2022.
\newblock QueryProp: Object Query Propagation for High-Performance Video Object
  Detection.
\newblock In \emph{AAAI}, 834--842.

\bibitem[{He et~al.(2017)He, Gkioxari, Doll{\'a}r, and Girshick}]{he2017mask}
He, K.; Gkioxari, G.; Doll{\'a}r, P.; and Girshick, R. 2017.
\newblock Mask R-CNN.
\newblock In \emph{ICCV}, 2961--2969.

\bibitem[{He et~al.(2016)He, Zhang, Ren, and Sun}]{he2016deep}
He, K.; Zhang, X.; Ren, S.; and Sun, J. 2016.
\newblock Deep residual learning for image recognition.
\newblock In \emph{CVPR}, 770--778.

\bibitem[{He et~al.(2021)He, Zhou, Li, Niu, Cheng, Li, Liu, Tong, Ma, and
  Zhang}]{he2021end}
He, L.; Zhou, Q.; Li, X.; Niu, L.; Cheng, G.; Li, X.; Liu, W.; Tong, Y.; Ma,
  L.; and Zhang, L. 2021.
\newblock End-to-End Video Object Detection with Spatial-Temporal Transformers.
\newblock In \emph{ACM MM}, 1507--1516.

\bibitem[{Hu et~al.(2018)Hu, Gu, Zhang, Dai, and Wei}]{hu2018relation}
Hu, H.; Gu, J.; Zhang, Z.; Dai, J.; and Wei, Y. 2018.
\newblock Relation networks for object detection.
\newblock In \emph{CVPR}, 3588--3597.

\bibitem[{Krizhevsky, Sutskever, and Hinton(2012)}]{krizhevsky2012imagenet}
Krizhevsky, A.; Sutskever, I.; and Hinton, G.~E. 2012.
\newblock Imagenet classification with deep convolutional neural networks.
\newblock In \emph{NIPS}, 1097--1105.

\bibitem[{Lin et~al.(2017{\natexlab{a}})Lin, Doll{\'a}r, Girshick, He,
  Hariharan, and Belongie}]{lin2017feature}
Lin, T.-Y.; Doll{\'a}r, P.; Girshick, R.; He, K.; Hariharan, B.; and Belongie,
  S. 2017{\natexlab{a}}.
\newblock Feature pyramid networks for object detection.
\newblock In \emph{CVPR}, 2117--2125.

\bibitem[{Lin et~al.(2017{\natexlab{b}})Lin, Goyal, Girshick, He, and
  Doll{\'a}r}]{lin2017focal}
Lin, T.-Y.; Goyal, P.; Girshick, R.; He, K.; and Doll{\'a}r, P.
  2017{\natexlab{b}}.
\newblock Focal loss for dense object detection.
\newblock In \emph{ICCV}, 2980--2988.

\bibitem[{Lin et~al.(2014)Lin, Maire, Belongie, Hays, Perona, Ramanan,
  Doll{\'a}r, and Zitnick}]{lin2014microsoft}
Lin, T.-Y.; Maire, M.; Belongie, S.; Hays, J.; Perona, P.; Ramanan, D.;
  Doll{\'a}r, P.; and Zitnick, C.~L. 2014.
\newblock Microsoft coco: Common objects in context.
\newblock In \emph{ECCV}, 740--755.

\bibitem[{Liu et~al.(2018)Liu, Qi, Qin, Shi, and Jia}]{liu2018path}
Liu, S.; Qi, L.; Qin, H.; Shi, J.; and Jia, J. 2018.
\newblock Path aggregation network for instance segmentation.
\newblock In \emph{CVPR}, 8759--8768.

\bibitem[{Liu et~al.(2016)Liu, Anguelov, Erhan, Szegedy, Reed, Fu, and
  Berg}]{liu2016ssd}
Liu, W.; Anguelov, D.; Erhan, D.; Szegedy, C.; Reed, S.; Fu, C.-Y.; and Berg,
  A.~C. 2016.
\newblock SSD: Single Shot Multibox Detector.
\newblock In \emph{ECCV}, 21--37.

\bibitem[{Redmon et~al.(2016)Redmon, Divvala, Girshick, and
  Farhadi}]{redmon2016you}
Redmon, J.; Divvala, S.; Girshick, R.; and Farhadi, A. 2016.
\newblock You only look once: Unified, real-time object detection.
\newblock In \emph{CVPR}, 779--788.

\bibitem[{Redmon and Farhadi(2017)}]{redmon2017yolo9000}
Redmon, J.; and Farhadi, A. 2017.
\newblock YOLO9000: better, faster, stronger.
\newblock In \emph{CVPR}, 7263--7271.

\bibitem[{Ren et~al.(2015)Ren, He, Girshick, and Sun}]{ren2015faster}
Ren, S.; He, K.; Girshick, R.; and Sun, J. 2015.
\newblock Faster R-CNN: Towards real-time object detection with region proposal
  networks.
\newblock In \emph{NIPS}, 91--99.

\bibitem[{Russakovsky et~al.(2015)Russakovsky, Deng, Su, Krause, Satheesh, Ma,
  Huang, Karpathy, Khosla, Bernstein et~al.}]{russakovsky2015imagenet}
Russakovsky, O.; Deng, J.; Su, H.; Krause, J.; Satheesh, S.; Ma, S.; Huang, Z.;
  Karpathy, A.; Khosla, A.; Bernstein, M.; et~al. 2015.
\newblock Imagenet large scale visual recognition challenge.
\newblock \emph{IJCV}, 211--252.

\bibitem[{Sabater, Montesano, and Murillo(2020)}]{sabater2020robust}
Sabater, A.; Montesano, L.; and Murillo, A.~C. 2020.
\newblock Robust and efficient post-processing for video object detection.
\newblock In \emph{IROS}, 10536--10542.

\bibitem[{Shvets, Liu, and Berg(2019)}]{shvets2019leveraging}
Shvets, M.; Liu, W.; and Berg, A.~C. 2019.
\newblock Leveraging long-range temporal relationships between proposals for
  video object detection.
\newblock In \emph{ICCV}, 9756--9764.

\bibitem[{Simonyan and Zisserman(2015)}]{simonyan2014very}
Simonyan, K.; and Zisserman, A. 2015.
\newblock Very deep convolutional networks for large-scale image recognition.
\newblock In \emph{ICLR}.

\bibitem[{Sun et~al.(2021)Sun, Hua, Hu, and Robertson}]{sun2021mamba}
Sun, G.; Hua, Y.; Hu, G.; and Robertson, N. 2021.
\newblock MAMBA: Multi-level Aggregation via Memory Bank for Video Object
  Detection.
\newblock In \emph{AAAI}, 2620--2627.

\bibitem[{Tian et~al.(2019)Tian, Shen, Chen, and He}]{tian2019fcos}
Tian, Z.; Shen, C.; Chen, H.; and He, T. 2019.
\newblock Fcos: Fully convolutional one-stage object detection.
\newblock In \emph{ICCV}, 9627--9636.

\bibitem[{Vaswani et~al.(2017)Vaswani, Shazeer, Parmar, Uszkoreit, Jones,
  Gomez, Kaiser, and Polosukhin}]{vaswani2017attention}
Vaswani, A.; Shazeer, N.; Parmar, N.; Uszkoreit, J.; Jones, L.; Gomez, A.~N.;
  Kaiser, {\L}.; and Polosukhin, I. 2017.
\newblock Attention is all you need.
\newblock In \emph{NIPS}, 5998--6008.

\bibitem[{Wang et~al.(2020)Wang, Liao, Wu, Chen, Hsieh, and
  Yeh}]{wang2020cspnet}
Wang, C.-Y.; Liao, H.-Y.~M.; Wu, Y.-H.; Chen, P.-Y.; Hsieh, J.-W.; and Yeh,
  I.-H. 2020.
\newblock CSPNet: A new backbone that can enhance learning capability of CNN.
\newblock In \emph{CVPR}, 390--391.

\bibitem[{Wu et~al.(2019)Wu, Chen, Wang, and Zhang}]{wu2019sequence}
Wu, H.; Chen, Y.; Wang, N.; and Zhang, Z. 2019.
\newblock Sequence level semantics aggregation for video object detection.
\newblock In \emph{ICCV}, 9217--9225.

\bibitem[{Xie et~al.(2017)Xie, Girshick, Doll{\'a}r, Tu, and
  He}]{xie2017aggregated}
Xie, S.; Girshick, R.; Doll{\'a}r, P.; Tu, Z.; and He, K. 2017.
\newblock Aggregated residual transformations for deep neural networks.
\newblock In \emph{CVPR}, 1492--1500.

\bibitem[{Xu et~al.(2022)Xu, Wang, Lv, Chang, Cui, Deng, Wang, Dang, Wei, Du
  et~al.}]{xu2022pp}
Xu, S.; Wang, X.; Lv, W.; Chang, Q.; Cui, C.; Deng, K.; Wang, G.; Dang, Q.;
  Wei, S.; Du, Y.; et~al. 2022.
\newblock PP-YOLOE: An evolved version of YOLO.
\newblock \emph{arXiv preprint arXiv:2203.16250}.

\bibitem[{Zhang et~al.(2018{\natexlab{a}})Zhang, Cisse, Dauphin, and
  Lopez-Paz}]{zhang2018mixup}
Zhang, H.; Cisse, M.; Dauphin, Y.~N.; and Lopez-Paz, D. 2018{\natexlab{a}}.
\newblock mixup: Beyond Empirical Risk Minimization.
\newblock In \emph{ICLR}.

\bibitem[{Zhang et~al.(2018{\natexlab{b}})Zhang, Cheng, Zhu, Lin, and
  Dai}]{zhang2018integrated}
Zhang, Z.; Cheng, D.; Zhu, X.; Lin, S.; and Dai, J. 2018{\natexlab{b}}.
\newblock Integrated object detection and tracking with tracklet-conditioned
  detection.
\newblock \emph{arXiv preprint arXiv:1811.11167}.

\bibitem[{Zhu et~al.(2018)Zhu, Dai, Yuan, and Wei}]{zhu2018towards}
Zhu, X.; Dai, J.; Yuan, L.; and Wei, Y. 2018.
\newblock Towards high performance video object detection.
\newblock In \emph{CVPR}, 7210--7218.

\bibitem[{Zhu et~al.(2017{\natexlab{a}})Zhu, Wang, Dai, Yuan, and
  Wei}]{zhu2017flow}
Zhu, X.; Wang, Y.; Dai, J.; Yuan, L.; and Wei, Y. 2017{\natexlab{a}}.
\newblock Flow-guided feature aggregation for video object detection.
\newblock In \emph{ICCV}, 408--417.

\bibitem[{Zhu et~al.(2017{\natexlab{b}})Zhu, Xiong, Dai, Yuan, and
  Wei}]{zhu2017deep}
Zhu, X.; Xiong, Y.; Dai, J.; Yuan, L.; and Wei, Y. 2017{\natexlab{b}}.
\newblock Deep feature flow for video recognition.
\newblock In \emph{CVPR}, 2349--2358.

\end{thebibliography}
\end{document}